\documentclass[12pt]{article}
\usepackage[margin=1in]{geometry}
\usepackage{graphicx,url}
\usepackage[utf8]{inputenc}
\usepackage{amsmath, amssymb, amsthm}
\usepackage[ruled]{algorithm2e}
\usepackage{caption}
\usepackage{algpseudocode}
\usepackage[english]{babel}
\usepackage{xcolor,soul}
\usepackage{natbib}
\usepackage{hyperref}
\usepackage{bbm,bm}
\usepackage{xr,tcolorbox}

\def\bb{\bm b}
   
  \def\bbeta{\bm \beta}
  \def\bdelta{\bm \delta}
  \def\bw{\bm w}

  \def\bSig{\bm \Sigma}
  \def\bH{\bm H}
  \def \bx{\bm x}
  \def \bX{\bm X}
  \def \by{\bm y}
  \def\P{\mathbb{P}}
  \def\R{\mathbb{R}}
  \def\E{\mathbb{E}}

  \def\lam{\lambda}

  \def\mN{\mathcal{N}}

  \def\btheta{\bm{\theta}}
  
  \usepackage{mathtools}
  
  \DeclareMathOperator*{\argmax}{arg\,max}
\DeclareMathOperator*{\argmin}{arg\,min}
\newtheorem{theorem}{Theorem}[section]
\newtheorem{lemma}{Lemma}[section]

\newtheorem{corollary}{Corollary}[section]
\newtheorem{remark}{Remark}[section]
\newtheorem{condition}{Condition}[section]
\usepackage{chngcntr}
\definecolor{darkred}{RGB}{150,50,50}
\definecolor{brown}{RGB}{250,100,100}
\definecolor{green}{RGB}{000,150,100}
\definecolor{purple}{RGB}{250,000,180}

\usepackage{hyperref}
\let\svthefootnote\thefootnote
\newcommand\freefootnote[1]{%
	\let\thefootnote\relax%
	\footnotetext{#1}%
	\let\thefootnote\svthefootnote%
}

\title{Targeting Underrepresented Populations in Precision Medicine: A Federated Transfer Learning Approach}
\author{Sai Li$^1$, Tianxi Cai$^2$, and Rui Duan$^2$$^*$\\
	\\
$^1$Institute of Statistics and Big Data, Renmin University of China\\
$^2$Department of Biostatistics, Harvard University}
\counterwithin{figure}{section}
\begin{document} 

\maketitle
\begin{abstract}
The limited representation of minorities and disadvantaged populations in large-scale clinical and genomics research has
 become a barrier to translating precision medicine research into practice. Due to heterogeneity across populations, risk prediction models are often found to be underperformed in these underrepresented populations, and therefore may further exacerbate known health disparities. 
In this paper, we propose a two-way data integration strategy that integrates heterogeneous data from diverse populations and from multiple healthcare institutions via a federated transfer learning approach. The proposed method can handle the challenging setting where sample sizes from different populations are highly unbalanced. With only a small number of communications across participating sites, the proposed method can achieve performance comparable to the pooled analysis where individual-level data are directly pooled together. We show that the proposed method improves the estimation and prediction accuracy in underrepresented populations, and reduces the gap of model performance across populations.  Our theoretical analysis reveals how estimation accuracy is influenced by communication budgets, privacy restrictions, and heterogeneity across populations. We demonstrate the feasibility and validity of our methods through numerical experiments and a real application to a multi-center study, in which we construct polygenic risk prediction models for Type II diabetes in AA population. 
\end{abstract}
\freefootnote{* Corresponding to rduan@hsph.harvard.edu}
\section{Introduction}
\subsection{Motivation}
Personalized medicine holds promises to improve individual health by integrating a person's genetics, environment, and lifestyle information to determine the best approach to prevent or treat diseases \citep{ashley2016towards}. Precision medicine research has attracted considerable interest and investment during the past few decades \citep{collins2015new}.  With the emergence of electronic health records (EHR) linked with biobank specimens, massive environmental data, and health surveys,  we now have increasing opportunities to develop accurate personalized risks prediction models  in a cost-effective way \citep{li2020electronic}. 

 Despite the availability of large-scale biomedical data, many demographic sub-populations are observed to be underrepresented in precision medicine research \citep{west2017genomics,kraft2018beyond}. For example, a disproportionate majority ($>$75\%) of participants in existing genomics studies are of European descent \citep{martin2019clinical}. The UK biobank, one of the largest biobanks, has more than 95\% of European-ancestry (EA) participants \citep{sudlow2015uk}. It remains challenging to optimize prediction model performance for such underrepresented  populations, when there is a substantial amount of heterogeneity in underlying distributions of data across populations \citep{west2017genomics,landry2018lack,kraft2018beyond,duncan2019analysis}. For some diseases, due to the differences in genetic architectures, linkage disequilibrium (LD) structures, and minor allele frequencies  across ancestral populations,   the performance of genetic risk prediction models in non-European populations has generally been found to be much poorer than in EA populations, most notably in African ancestry (AA) populations \citep{duncan2019analysis}. To advance prediction medicine, it is crucial to improve the performance of statistical and machine learning models in underrepresented populations so as not to exacerbate health disparities.

We proposed to address the lack of representation and disparities in model performance through two data integration strategies: (1)  leveraging the shared knowledge from diverse populations, and (2) integrate larger bodies of data from multiple healthcare institutions. 
Data across multiple populations may share a certain amount of similarity that can be leveraged to improve the model performance in an underrepresented population \citep{cai2021unified}. However, conventional methods where all data are combined and used indistinctly in training and testing, cannot tailor the prediction models to work well for specific population \citep{duncan2019analysis}. To account for such heterogeneity and lack of representation, we propose to use transfer learning to transfer the shared knowledge learned from diverse populations to an underrepresented population, so that comparable model performance can be reached with much less data for training \citep{weiss2016survey}.  In addition, multi-institutional data integration can improve the sample size of the underrepresented populations and the diversity of data \citep{mccarty2011emerge}. We propose to use federated learning to unlock the multi-institutional EHR/biobank data, which overcomes two main barriers of institutional data integration. One is that the individual-level information can be highly sensitive which cannot be shared across institutions \citep{van2003data}. The other is that the often-enormous size of the EHR/biobanks data makes it infeasible or inefficient to pooling all data together due to challenges in data storage, management, and computation \citep{kushida2012strategies}. Therefore, as illustrated in Figure \ref{fig1}, our goal is to develop a federated transfer framework to incorporate data from diverse populations that are stored at multiple institutions to improve the model performance in a target underrepresented population.

\begin{figure}
\caption{A schematic illustration of the federated transfer learning framework and the problem setting.}
\centering
\includegraphics[width= 15 cm]{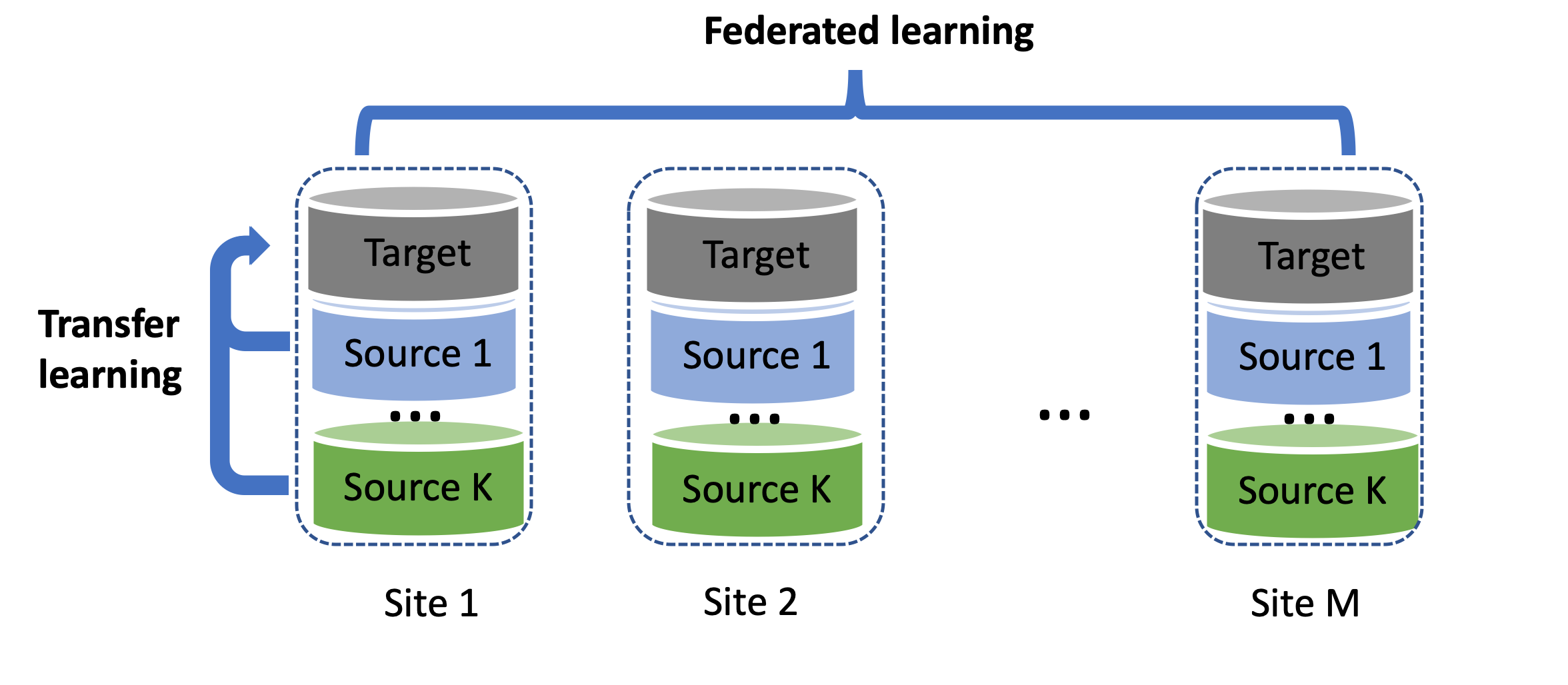}
\end{figure}\label{fig1}

\subsection{Related Work}

Existing transfer learning methods primarily focus on settings where individual-level data can be shared. For example, \citet{CW19} studied the minimax and adaptive methods for nonparametric classification in the transfer learning setting. \citet{Bastani18} studied estimation and prediction in high-dimensional linear models and the sample size of the auxiliary study is larger than the number of covariates. \citet{li2020transfer} propose a minimax optimal transfer learning algorithm in high-dimensional linear models and study the adaptation to the unknown similarity level. \citet{Li2021GGM} studies transfer learning in high-dimensional Gaussian graphical models with false discovery rate control. {\citet{LZCL21} studies transfer learning in high-dimensional generalized linear models (GLMs) and establishes the minimax optimality. \citet{tian2021transfer} studies adaptation to the unknown similarity level in transfer learning under high-dimensional GLMs. These individual data-based methods cannot be directly extended to the federated setting due to data sharing constraints and the potential heterogeneity across sites.


On the other hand, under data sharing constraints,  most federated learning methods focus on settings where the true models are the same across studies. 
For example, many algorithms fit a common model to data from each institution and then aggregate these local estimates through a weighted average \citep[e.g]{li2013statistical,chen2014split,lee2017communication, tian2016communication, lian2017divide, battey2018distributed,wang2019distributed}. 
To improve efficiency, surrogate likelihood approaches have  been adopted in recently proposed distributed algorithms \citep{jordan2018communication,duan2019odal,duan2020learningb}  to approximate the global likelihood. These methods cannot be easily extended to the federated transfer learning setting where both data sharing constraints and heterogeneity are present. 


{ 
Recently, \cite{liu2020integrative} and \citet{Xia21} proposed distributed multi-task learning approaches that account for both study heterogeneity and data privacy. They allow site-specific regression parameters that are assumed to be similar across sites in magnitude and support and perform integrative analyses based on derived summary data.  However, these methods require the sample sizes for different populations to be of the same order and can perform poorly when the underlying models of some source sites are significantly different from the target sites. Different from their work, we consider a more general setting where data from multiple populations are stored in multiple sites and a more challenging setting where the sample size from each population can be highly unbalanced. We focus on the model performance of an underrepresented target population without making assumptions that model parameters across populations share the support and magnitude. Instead, our methods are robust to the cases where the underlying models for some populations differ significantly from the target population.} 


\subsection{Contributions and main results}
We propose a methodology framework that incorporates heterogeneous data from diverse populations and multiple healthcare organizations to improve model fitting and prediction in an underrepresented population. Adopting transfer learning ideas, our methods tackle an important issue in precision medicine where sample sizes from different populations can be highly unbalanced. Our theoretical analysis and numerical experiments show that our methods are more accurate compared to existing methods and are robust to the level of heterogeneity. The federated learning methods we propose only require a small number of communications across participating sites, and can achieve performance comparable to the pooled analysis where individual-level data are directly pooled together. To the best of our knowledge, this is the first work that tailors transfer learning and federated computing towards improving the performance of models in underrepresented populations. From a high level, our theoretical analysis shows that the proposed methods reduce the gap of estimation accuracy across populations, and reveals how estimation accuracy is influenced by communication budgets, privacy restrictions, and heterogeneity among populations.  We demonstrate the feasibility and validity of our methods through numerical experiments and a real application to a multi-center study, in which we construct polygenic risk prediction models for Type II diabetes in AA population. 

\section{Method}
\subsection{Problem set-up and notation}
We build our federated transfer learning methods based on sparse high-dimensional regression models \citep{tibshirani1996regression, bickel2009simultaneous}. These models have been widely applied to precision medicine research for both association studies and risk prediction models, due to the benefits of simultaneous model estimation and variable selection, and the desirable interpretability \citep{qian2020fast}.

We assume there are $N$ subjects in total from  $K+1$ populations. We treat the underrepresented population of interest as the target population, indexed by $k=0$, while the other $K$ populations are treated as source populations, indexed by $k = 1, \dots K$. We assume data for the $N$ subjects are stored at $M$ different sites, where due to privacy constraints, no individual-level data are allowed to be shared across sites. {We consider the case where $K$ is finite but $M$ is allowed to grow as the total sample size grows to infinity.}

 Let $\mN^{(m,k)}$ be the index sets of the data from the $k$-th  population in the $m$-th site, and $n^{(m, k)}=|\mN^{(m,k)}|$ denote the corresponding sample size, for $k=0,\dots,K$ and $m  = 1, \dots M$.  We assume the index sets are known and do not overlap with one another, i.e., $\mN^{(m,k)}\cap \mN^{(m,k')}=\emptyset$ for any $k, k'\in{0, \dots, K}$, and $k\ne k'$. In precision medicine research, these index sets may be obtained from indicators of minority and disadvantaged groups, such as race/ethnics, gender, and socioeconomic status. Denote $N^{(k)}=\sum_{m=1}^Mn^{(m,k)}$, and $N=\sum_{k=0}^KN^{(k)}$.  We are particularly interested in the challenging scenario $N^{(0)}\ll N$, where the underrepresentation is severe. However, at certain sites,  the relative sample compositions can be arbitrary. It is possible that some sites may not have data from certain populations, i.e., $n^{(m,k)} =0 $ for  some but not all $m$ for $1\leq m\leq M$. We consider the high-dimensional setting where $p$ can be larger and much larger than $N^{(0)}$ and $N$.

\def\trans{^{\intercal}}

For the $i$-th subject, we observe an outcome variable $y_i\in\R$ and a set of $p$ predictors $\bx_i\in\R^p$ including the intercept term. We assume that the target data on the $m$-th site, $\{\bx_i,y_i\}_{i\in \mN^{(m,0)}}$,  follow a GLM 
\[
g \{\E(y_i|\bx_i)\} = \bx_i^{\intercal}\bbeta,
\]
with a canonical link function $g(\cdot)$ and a negative log-likelihood function
\[
L^{(m,0)}(\bbeta)=\sum_{i\in\mN^{(m,0)}}\{\psi(\bx_i^{\intercal}\bbeta)-y_i\cdot \bx_i^{\intercal}\bbeta\},
\]
for some unknown parameter $\bbeta\in\R^p$ and $\psi(\cdot)$ uniquely determined by $g(\cdot)$.
Similarly, the data from the $k$-th source population in the $m$-th site are $\{\bx_i,y_i\}_{i\in \mN^{(m,k)}}$ and they follow a GLM 
\[
g \{\E(y_i|\bx_i)\} = \bx_i^{\intercal}\bw^{(k)},
\]
with negative log-likelihood
\[
L^{(m,k)}(\bw^{(k)})=\sum_{i\in\mN^{(m,k)}}\{\psi(\bx_i^{\intercal}\bw^{(k)})-y_i\cdot \bx_i^{\intercal}\bw^{(k)}\}
\]
for some unknown parameter $\bw^{(k)}\in\R^p$.  

Our goal is to estimate $\bbeta$, using data from  the $K+1$  populations from the $M$ sites. These data are heterogeneous at two levels:  For data from different populations, differences may exist in terms of both the regression coefficients, which characterize conditional distribution $f(y_i|\bx_i)$, as well as the underlying distribution of the covariates $f(\bx_i)$,  also known as covariate shift in some related work \citep{guo2020inference}. For data from a given population, the distribution of covariates $f(\bx_i)$ might also be heterogeneous across sites. In addition to the heterogeneity, we consider the setting only summary-level data can be shared across sites. Thus, we assume the regression parameters to be distinct across populations and, given a specific population, the regression parameter is the same across sites.


{
Despite the presence of between-population heterogeneity,  it is reasonable to believe that the population-specific models share some degree of similarity. For example, the genetic architectures, captured by regression coefficients,  of many complex traits and diseases are found to be highly concordant across ancestral groups \citep{lam2019comparative}. It is important to characterize and leverage such similarities so that knowledge can be transferred from the source to the target population. 

Under our proposed modeling framework, we characterize the similarities between the $k$-th source population and the target based on the difference between their  regression parameters, $\bdelta^{(k)} = \bw^{(k)} - \bbeta$.  We consider the following parameter space
\[
\Theta(s,h)=\left\{\btheta = (\bbeta,\bdelta^{(1)}, \dots, \bdelta^{(K)}): \|\bbeta\|_0\le s, \max_{1\leq k\leq K}\|\bdelta^{(k)}\|_0\le h\right\},
\]
where $s$ and $h$ are the upper bounds for the support size of $\bbeta$ and $\{\bdelta^{(k)}\}_{k=1}^K$, respectively. Intuitively,  smaller  $h$ indicates a higher level of  similarity, so that the source data can be more helpful for estimating $\bbeta$ in the target population. When $h$ is relatively large, incorporating data from  source populations may be worse than only using data from the target population to fit the model, also known as negative transfer in the machine learning literature \citep{weiss2016survey}. With unknown $s$ and $h$ in practice, we aim to devise an adaptive estimator to avoid negative transfer under unknown levels of heterogeneity across populations.
}

\subsection{The proposed algorithm}
\label{sec2-2}

Throughout, for real-valued sequences $\{a_n\}, \{b_n\}$, we write $a_n \lesssim b_n$ if $a_n \leq cb_n$ for some universal constant $c \in (0, \infty)$, and $a_n \gtrsim b_n$ if $a_n \geq c'b_n$ for some universal constant $c' \in (0, \infty)$.  We say $a_n \asymp b_n$ if $a_n \lesssim b_n$ and $a_n \gtrsim b_n$. We let $c, C, c_0, c_1, c_2, \cdots, $ denote some universal constants. 
For a vector $\bm v \in \R^d$ and an index set $S \subseteq [d]$, we use $\bm v_S$ to denote the subvector of $\bm v$ corresponding to $S$. For any vector $\bb\in\R^p$, let $\mathcal{H}_k(\bb)$ be formed by setting all but the largest (in magnitude) $k$ elements of $\bb$ to zero. For a matrix $A\in\R^{n_1\times n_2}$, let $\Lambda_{\max} (A)$ and $\Lambda_{\min}(A)$  denote the largest and smallest singular values of $A$,  and $\|A\|_{\infty,\infty}\max_{i\leq n_1,j\leq n_2}|A_{i,j}|$. For a random variable $u\in\R$ and a random vector $\bm U\in\R^{n}$, define their sub-Gaussian norms as $\|u\|_{\psi_2}=\sup_{l\geq 1} l^{-1/2}\E^{1/l}[|u|^l]$ and
$\|\bm U\|_{\psi_2}=\sup_{\|\bm v\|_2=1,\bm v\in\R^{n}}\|\langle \bm U,\bm v\rangle\|_{\psi_2}$. 

To motivate our proposed federated transfer learning algorithm, we first consider the ideal case when site-level data can be shared. The transfer learning estimator of $\bbeta$ under the high-dimensional GLMs can be obtained via the following three-step procedure:

\noindent\underline{Step 1:} Fit a regression model  in each source population. For $k \in \{1, \dots, K\}$,  we obtain 
\begin{align}
\hat{\bw}^{(k)}&=\argmin_{\bb\in\R^p} \left\{\frac{1}{N^{(k)}}\sum_{m=1}^ML^{(m,k)}(\bb)+\lam^{(k)}\|\bb\|_1\right\}\label{global1}
\end{align}
\underline{Step 2:} Adjust for differences using target data. For $k=1,\dots,K$, we obtain 
\begin{align}
\hat{\bdelta}^{(k)}&=\argmin_{\bb\in\R^p} \left\{\frac{1}{N^{(0)}}\sum_{m=1}^ML^{(m,0)}(\hat{\bw}^{(k)}+\bb)+\lam_{\delta}\|\bb\|_1\right\}\label{global2}.
\end{align} 
Threshold $\hat{\bdelta}^{(k)}$ via $\check{\bdelta}^{(k)}=\mathcal{H}_{\sqrt{N^{(0)}/\log p}}(\hat{\bdelta}^{(k)})$.

\noindent\underline{Step 3:} Joint estimation using source and target data
\begin{align}
\hat{\bbeta}&=\argmin_{\bb\in\R^p} \left\{\frac{1}{N}\sum_{m=1}^ML^{(m,0)}(\bb)+\frac{1}{N} \sum_{k=1}^K\sum_{m=1}^ML^{(m,k)}(\bb-\check{\bdelta}^{(k)}) +\lam_{\beta}\|\bb\|_1\right\},\label{global3}
\end{align} 
where $\{\lam^{(k)}\}_{k=1}^K$, $\lam_{\delta}$, and $\lam_{\beta}$ are tuning parameters. Instead of learning $\bbeta$ directly from the target data which have limited sample size, we learn $\bw^{(k)}$ from the source populations, and use them to ``jumpstart" the model fitting in the target population. More specifically, we learn the difference $\bdelta^{(k)}$ by offsetting each $\hat\bw^{(k)}$. In Step 3, we combine all the data together to jointly learn $\bbeta$, where the estimated differences  $\hat\bdelta^{(k)}$ are adjusted for data from the $k$-th source population. 
In contrast to existing transfer learning methods based on GLM, the above procedure has benefits in estimation accuracy and flexibility to be implemented in the federated setting. Compared to a recent work \citep{tian2021transfer}, the above procedure has a faster convergence rate, which is in fact minimax optimal under mild conditions. Moreover, our method learns $\bw^{(k)}$ independently in Step 1 and Step 2, while in other related methods \citep{tian2021transfer,li2020transfer},  a pooled analysis is conducted with data from multiple populations. In a federated setting, finding a proper initialization is challenging for such a pooled estimator due to various levels of heterogeneity. In addition, compared to \cite{tian2021transfer}, the above approach has fewer assumptions on the level of heterogeneity for data from different populations.

To generalize (\ref{global1})-(\ref{global3}) to the federated setting, we consider an approximation of $L^{(m,k)}(\bb)$ by its the second-order expansion of $L^{(m,k)}(\bb)$ at $\mathring{\bb}$. That is,  
\begin{align*}
\tilde L^{(m,k)}(\bb;\mathring{\bb})&= L^{(m,k)}(\mathring{\bb})+ \sum_{m=1}^M( \bb-\mathring{\bb})\trans\nabla{L}^{(m,k)}(\mathring{\bb})+\frac{1}{2}\sum_{m=1}^M\nabla^2{L}^{(m,k)}(\mathring{\bb})(\bb-\mathring{\bb})^{\otimes 2}. 
\end{align*}
 The higher-order terms are omitted given that the initial value $\mathring{\bb}$ is sufficiently close to the true parameter.
Using these surrogate losses, the sites only need to share three sets of summary statistics, $\mathring{\bb}$, the score vector $\nabla{L}^{(m,k)}(\mathring{\bb})$ and the Hessian matrix $\nabla^2{L}^{(m,k)}(\mathring{\bb})$.  For $k=0,\dots,K$, we define 
\begin{align}
&\nabla L^{(k)}(\bb)=\sum_{m=1}^{M}\nabla L^{(m,k)}(\bb), ~\widehat{\bH}^{(k)}(\bb)=\frac{1}{N^{(k)}}\sum_{m=1}^M\nabla^2{L}^{(m,k)}(\bb),\nonumber\\
&R^{(k)}(\bb;\mathring{\bb})=\frac{1}{2}(\bb-\mathring{\bb})^{\intercal}\widehat{\bH}^{(k)}(\mathring{\bb})(\bb-\mathring{\bb})+\langle \bb-\mathring{\bb},\frac{1}{N^{(k)}}\nabla L^{(k)}(\mathring{\bb})\rangle.\label{eq-gradient}
\end{align}
The functions $\widehat{R}^{(k)}(\bb;\mathring{\bb})$ are the combined surrogate log-likelihood functions for the $k$-th population based on some previous estimate $\mathring{\bb}$ and corresponding gradients obtained from the $M$ sites. We then follow similar strategies as (\ref{global1})-(\ref{global3}) but replace the full likelihood with the surrogate losses to construct a federated transfer learning estimator for $\bbeta$, as detailed in  Algorithm \ref{alg-ms}.\\

\begin{algorithm}[H]
	\SetKwInOut{Input}{Input}
	\SetKwInOut{Output}{Output}
	\Input{Target population$\{\bX^{(m,0)},\by^{(m,0)}\}_{m=1}^M$ and source populations \{$\{\bX^{(m,k)},\by^{(m,k)}\}_{m=1}^M\}_{k=1}^K$. \\
	Initial values $\hat{\bbeta}_0$, $\{\hat{\bw}^{(k)}_0\}_{k=1}^K$.}
	\Output{$\hat{\bbeta}_T$}
	
	
	\For{$t=1,\dots,T$}{
			Threshold   $\check{\bw}_{t-1}^{(k)}=\mathcal{H}_{c_n}(\hat{\bw}^{(k)}_{t-1})$ and $\check{\bbeta}_{t-1}=\mathcal{H}_{c_n}(\hat{\bbeta}_{t-1})$.\\
		\For{$m=1,\dots, M$}{ 
			Transmit $\{\nabla L^{(m,0)}(\check{\bbeta}_{t-1}),\{\nabla L^{(m,k)}(\check{\bw}_{t-1}^{(k)})\}_{k=1}^K\}$ and $\{\nabla^2L^{(m,0)}(\check{\bbeta}_{t-1}),\{\nabla^2 L^{(m,k)}(\check{\bw}_{t-1}^{(k)})\}_{k=1}^K\}$ to the leading site.
		}
		\textbf{Compute} 
		the combined first- and second-order information  $\nabla L^{(0)}(\check{\bbeta}_{t-1})$,  $\nabla L^{(k)}(\check{\bw}_{t-1}^{(k)})$, 
		$\widehat{\bH}^{(0)}(\check{\bbeta}_{t-1})$,  and $\widehat{\bH}^{(k)}(\check{\bw}_{t-1}^{(k)})$ according to (\ref{eq-gradient}). Compute
		\begin{align}
		&\hat{\bw}^{(k)}_t=\argmin_{\bb\in\R^p}\left\{\widehat{R}^{(k)}(\bb;\check{\bw}^{(k)}_{t-1})+\lam^{(k)}\|\bb\|_1\right\}, ~k=1,\dots,K.\label{est1}\\
		&\hat{\bdelta}_t^{(k)}=\argmin_{\bdelta\in\R^p}\left\{\widehat{R}^{(0)}(\hat{\bw}^{(k)}_t+\bdelta; \check{\bbeta}_{t-1})+\lam_{\delta}\|\bdelta\|_1\right\}, ~k=1,\dots,K.\label{est2}
		\end{align}
		Let $\check{\bdelta}_t^{(k)}=\mathcal{H}_{\sqrt{N^{(0)}/\log p}}(\hat{\bdelta}_t^{(k)})$, $k=1,\dots,K$.
		
		\textbf{Combine all the populations}:
		\begin{align}
		&\hat{\bbeta}_t=\argmin_{\bb\in\R^p}\left\{ \frac{N^{(0)}}{N}\widehat{R}^{(0)}(\bb;\check{\bbeta}_{t-1})+\sum_{k=1}^K\frac{N^{(k)}}{N}\widehat{R}^{(k)}(\bb+\check{\bdelta}_t^{(k)};\check{\bbeta}_{t-1})+\lam_{\beta}\|\bb\|_1\right\}.\label{joint-est2}
		\end{align} 
	}
	\caption{Federated transfer learning}
	\label{alg-ms}
\end{algorithm}
\begin{remark}
We discuss strategies for the initialization of $\bbeta$ and $\{\bw^{(k)}\}_{k = 1}^K$ in Section \ref{sec2-3}. Algorithm \ref{alg-ms} requires $T$ iterations, where within each iteration we collect the first- and second-order derivatives calculated at each site based on the current parameter values.  In practice, when iterative communication across sites is not preferred, we can choose $T=1$. We show in Section \ref{sec3} that additional iterations can improve the estimation accuracy. Proper choices of tuning parameters are also discussed in the sequel. In practical implementation, they can be chosen by cross-validation. 
\end{remark}

{When the source models are substantially different from the target model, the learned estimator $\hat{\bbeta}_T$ may not be better than a target only estimator, which is obtained using only the target data.} 
We thus proposed to increase the robustness of the transfer learning by optimally combining $\hat{\bbeta}_T$ with a target only estimator. This step can guarantee that, loosely speaking, the aggregated estimator has prediction performance comparable to the best prediction performance among all the candidate estimators \citep{RT11, Tsybakov14, Qagg}.  To this end, let $\hat{\bbeta}^{(tar)}_T$ denote a federated target-only estimator, whose construction is detailed the Supplementary Material.  This procedure can be aligned  with Algorithm \ref{alg-ms} in the implementation to reduce number of communications. With  $\hat{\bbeta}_T$ and $\hat{\bbeta}^{(tar)}_T$, we perform aggregation using some additional validation data from the target population in a leading site (denoted as the $m^*$-th site), which should not have any overlap with the training data used for obtaining $\hat{\bbeta}_T$ and $\hat{\bbeta}^{(tar)}_T$. In the leading site, we denote the validation data to be $\{\mathring{y}_i, \mathring{\bx}_i\}_{\{i = 1, \dots, \mathring{n}\}}$, with  sample size $\mathring{n} = cn^{(m^*,0)}$ for some $c\in (0, 1)$, where $n^{(m^*,0)}$ is the sample size of the training data in the leading site from the target population. 
Define $\widehat{\bm B}=(\hat{\bbeta}_T^{(tar)},\hat{\bbeta}_T)\in\R^{p\times 2}$. We compute
\begin{align*}
\hat{\bm \eta}=\argmin_{\eta\in\{e_1,e_2\}}\left\{\sum_{i=1}^{\mathring{n}}\mathring{y}_i\cdot (\mathring{\bx}_i)^{\intercal}\widehat{\bm B}{\bm\eta} -\psi((\mathring{\bx}_i)^{\intercal}\widehat{\bm B}{\bm\eta})\right\}.
\end{align*}
And the proposed estimator is defined as
$
\hat{\bbeta}^{agg} = \widehat{\bm B}\hat{\bm \eta}. 
$
Based on our simulation study and real data example, the size of the validation data can be relatively small compared to the training data, and cross-fitting may be used to make full use of all the data.  In practice, if there are strong prior knowledge indicating that the level of heterogeneity is low across populations, the aggregation step may be skipped.


\begin{remark} {(Avoid sharing  Hessian matrices)}
Algorithm \ref{alg-ms} requires each site to transmit Hessian matrices to the leading site, which may not be a concern when  $p$ is relatively small.   When $p$ is large, we provide possible options to reduce  communication cost of sharing Hessian matrices: (1) If the distributions of covariate variables $\bx$ are homogeneous across site for a certain population,  we propose to use Algorithm \ref{alg-local}, which only requires  the first-order gradients from each site. 
(2) When the distributions of covariate variables are heterogeneous across sites, if it is possible to fit a density ratio model between each dataset and the leading target data, we can still use the leading target data to approximate the Hessian matrices of the other datasets, through the density ratio tilting technique proposed in \cite{duan2019heterogeneity}. 
(3) We can leverage the sparsity structures of the population-level Hessian matrices, denoted by $\bH^{(m,k)}$,  to reduce the communication cost. For example, when constructing polygenic risk prediction, the existing knowledge on LD structure may infer similar block-diagonal structures of the Hessian matrices. In such cases, we can apply thresholding to the Hessian matrices and only share the resulting blocks.
(4) As demonstrated in our simulation study and real data application, our algorithm with one round of iteration (T=1) already achieves comparable performance as the pooled analysis. Thus, if choosing $T=1$, each site will only need to share Hessian matrices once. If more iterations are allowed, we propose an alternative algorithm where only the first-order gradients are needed in the rest of the $T-1$ iterations. More details are included in the supplements. 
\end{remark}


\subsection{Leveraging local Hessian under design homogeneity}
\label{sec2-3}
When the distribution of $\bx$ in the $k$th population is the same across sites, we introduce a modified version of Algorithm \ref{alg-ms} which only requires each participating site sharing only the first-order gradients. This method generalizes the surrogate likelihood approach proposed by \citet{wang2017efficient,jordan2018communication} to the  transfer learning framework and it enjoys communication efficiency. The idea of this algorithm is to use the local data to approximate the Hessian matrices across multiple sites. We require that the leading site (the $m^*$-th site) has data from all the $(K+1)$ populations. We will use the empirical Hessian matrix obtained  at the leading site to the approximate of the global Hessian in each population. For $k=0,\dots,K$, denote 
\[
   R^{(local,k)}(\bb;\mathring{\bb})=\frac{1}{2}(\bb-\mathring{\bb})^{\intercal}\widehat{\bH}^{(m^*,k)}(\mathring{\bb})(\bb-\mathring{\bb})+\langle \bb-\mathring{\bb},\frac{1}{N^{(k)}}\nabla L^{(k)}(\mathring{\bb})\rangle,~\text{where} 
\]
where
\[
  \widehat{\bH}^{(m^*,k)}(\mathring{\bb})=\frac{1}{n^{(m^*,k)}}\nabla^2 L^{(m^*,k)}(\mathring{\bb})
\]
is the empirical Hessian for the $k$-th population at $\bb'$ based on the samples in the leading site.

\begin{algorithm}[H]
	\SetKwInOut{Input}{Input}
	\SetKwInOut{Output}{Output}
	\Input{Target population$\{\bX^{(m,0)},\by^{(m,0)}\}_{m=1}^M$ and source populations \{$\{\bX^{(m,k)},\by^{(m,k)}\}_{m=1}^M\}_{k=1}^K$. }
	
Initial values $\hat{\bbeta}_0$, $\{\hat{\bw}^{(k)}_0\}_{k=1}^K$.

	\Output{$\hat{\bbeta}_T$}

	\For{$t=1,\dots,T$}{
		Threshold   $\check{\bw}_{t-1}^{(k)}=\mathcal{H}_{c_n}(\hat{\bw}^{(k)}_{t-1})$ and $\check{\bbeta}_{t-1}=\mathcal{H}_{c_n}(\hat{\bbeta}_{t-1})$.\\
		\For{$m=1,\dots, M$}{ 
			Transmit $\nabla L^{(m,0)}(\check{\bbeta}_{t-1})$ and $\{\nabla L^{(m,k)}(\check{\bw}_{t-1}^{(k)})\}_{k=1}^K$ to the leading site.
		}
	\textbf{Compute} 
		the combined first-order information  $\nabla L^{(0)}(\check{\bbeta}_{t-1})$,  $\nabla L^{(k)}(\check{\bw}_{t-1}^{(k)})$ according to (\ref{eq-gradient}).
		
	In (\ref{est1}), (\ref{est2}), and (\ref{joint-est2}) of Algorithm \ref{alg-ms},
	we replace $\widehat{R}^{(k)}(\bb;\bb')$ with $\widehat{R}^{(local,k)}(\bb;\bb')$  and replace $\lam^{(k)},\lam_{\delta},\lam_{\beta}$ with $\lam^{(k)}_t,\lam^{(k)}_{\delta,t},\lam_{\beta,t}$, respectively.
}
	\caption{Federated transfer learning leveraging local Hessian}
	\label{alg-local}
\end{algorithm}

Without sharing the Hessian matrices, Algorithm \ref{alg-local}  largely reduces the communication cost. However, one limitation is that it requires the distribution $f(\bx_{i})$ in the $k$-th population are homogeneous across sites for any fixed $k$. Second, its reliable performance requires existence of a single site  which has relatively large samples from all $K+1$ populations. Otherwise, the local Hessian approximation can be inaccurate and lead to large estimation errors. In practice, however, such a desirable local site may not always exist. We provide a theoretical comparison in Section \ref{sec3} showing that larger $T$ might be needed in Algorithm \ref{alg-local} to achieve the same estimation accuracy compared to \ref{alg-ms}.

\subsection{Initialization strategies}
\label{sec2-4}

The initialization determines the sample size requirements as well as the number of iterations Algorithms \ref{alg-ms} and \ref{alg-local} need to reach a convergence. With data from more than one populations, one needs to balance the sample sizes and similarities across populations. 
	
	Here we offer two initialization strategies, namely the single-site initialization and the multi-site initialization.  The ideal scenario for initialization is that one site has relatively large sample sizes for all $K+1$ populations. In such a case, we initialize $\bw^{(k)}$ and $\bbeta$ using the single-site initialization. 	If we cannot find a site with enough data from all the $K+1$ populations, the multi-site initialization can be used.
	
    \underline{Strategy 1: single-site initialization.}  Find  $m^* \in \{1, \dots M\}$, such that $n^{(m^*, k)} \asymp \max_{1\leq m\leq M} n^{(m, k)}$ for all $k \in \{0, \dots K\}$.  In site $m^*$,  we initialize $\bw^{(k)}$ and $\bbeta$ by applying the global transfer learning approach introduced in equations (\ref{global1})-(\ref{global3}). For example, the All of Us Precision Medicine Initiative aims to recruit 1 million Americans, with estimates of early recruitment showing up to 75\% of participants are from underrepresented populations. Such a dataset can be treated as an initialization site or leading site.

     \underline{Strategy 2: multi-site initialization.} 
     We first find $I_k = \argmax_{1\leq m\leq M} n^{(m,k)}$, which is the site with the largest sample size from the $k$-th population.  In site $I_k$, if sample size of the $k$-th populaiton is much smaller than the total sample size, we initialize $\bw^{(k)}$ by treating the $k$-th population as the target and other populations as the source, and apply the transfer learning approach introduced in equations (\ref{global1})-(\ref{global3}). If the $k$-th populaiton is the dominating population, we can simply initialize $\bw^{(k)}$ using only its own data. The same procedure applies to the initialization of  $\bbeta$.   For example, when constructing polygenic prediction models, the UK biobank has around 500k EA samples but only 3k AA samples. Thus, if UK biobank is selected for initialization of the EA-population, it can be done by using only the EA samples. On the contrary, if the UK biobank is selected for initialization of the AA population, a transfer learning approach is needed to improve the accuracy of initialization by incorporating both EA and AA samples.

     In Section \ref{sec3}, we show  that the convergence rates of Algorithm \ref{alg-ms} and Algorithm \ref{alg-local} depend on the accuracy of the initial estimators. For the above two strategies, we derive their corresponding convergence rates in the next section.

\section{Theoretical guarantees}
\label{sec3}

 Let $\bH^{(m,0)}=\E[\bx_i^{(m,0)}(\bx_i^{(m,0)})^{\intercal}\ddot{\psi}((\bx_i^{(m,0)})^{\intercal}\bbeta)]$ and $\bH^{(m,k)}=\E[\bx_i^{(m,k)}(\bx_i^{(m,k)})^{\intercal}\ddot{\psi}((\bx_i^{(m,k)})^{\intercal}\bw^{(k)})]$ denote the population Hessian matrices for the $k$-th population  at the $m$-th site. Let  $\bH^{(k)}=\sum_{m=1}^Mn^{(m,k)}\bH^{(m,k)}/N^{(k)}$, $k=0,\dots,K$, denote the population Hessian matrices for the $k$-th population across all sites. We assume the following condition for the theoretical analysis. 
\begin{condition}
	\label{cond1}
	For $m=1,\dots,M$, $k=0,\dots,K$,$\{\bx_i\}_{i\in\mN^{(m,k)}}$ are independent uniformly bounded with mean zero and covariance $\bSig^{(m,k)}$ with $\max_{1\leq m\leq M,0\leq k\leq K}\max_{i\in\in\mN^{(m,k)}}\|\bx_i\|_{\infty}\leq C<\infty$. 	The covariance matrices $\bSig^{(m,k)}$ and the Hessian matrices $\bH^{(m,k)}$ are all positive definite for $m=1,\dots,M$ and $k=1,\dots,K$. 
\end{condition}

\begin{condition}[Lipschitz condition of $\psi$]
	\label{cond2}
For $k=0,\dots,k$, the random noises $\{y_i-\dot{\psi}(\bx_i\bw^{(k)})\}_{i\in\mN^{(k)}}$ are independent sub-Gaussian with mean zero. The second-order derivative $\ddot{\psi}$ is uniformly bounded and $|\log\ddot{\psi}(a+b)-\log\ddot{\psi}(a)|\leq C|b|$ for any $a,b\in\R$.
\end{condition}

Condition \ref{cond1} assumes uniformly bounded designs with positive definite covariance matrices. The distribution of $\bx_i^{(m,k)}$ can be different for different $(m,k)$. This assumption is more realistic in biomedical setting than the homogeneity assumptions in, say, \citet{jordan2018communication}. In fact, the heterogeneous covariates are allowed because the Hessian matrices from different sites are transmitted in Algorithm \ref{alg-ms}. In contrast, Algorithm \ref{alg-local}, which  only require transmitting the gradients across sites, would require a stricter version of Condition \ref{cond1} as stated in Condition \ref{cond-homo}.
 On the other hand, the positive definiteness assumption is only required for the hessian matrices involved in the initialization, and the pooled Hessian matrix $\bH^{(k)}$. Moreover, when having unbounded covariates, one may consider relaxing the uniformly bounded designs to sub-Gaussian designs. We comment that our theoretical analysis still carry through with sub-Gaussian designs but the convergence rate will be inflated with some factors of $\log p$. Condition \ref{cond2} assume some standard Lipschitz conditions which hold for linear, logistic, and multinomial models. 

For the tuning parameters, we take 
	\[
	\lam^{(k)}=c_0\sqrt{\frac{\log p}{N^{(k)}}},~\lam_{\delta}=c_0\sqrt{\frac{\log p}{N^{(0)}}},~\lam_{\beta}=c_0\sqrt{\frac{\log p}{N}}+\frac{h\log p}{N^{(0)}}~\text{and}~c_n=2c_1s
	\]
for some constants $c_0>0$ and $c_1\geq 1$. We set $c_n$ at the magnitude of $s$ to simplify the theoretical analysis. In fact, $c_n$ depends on the sample size of the initialization site. With the initialization strategy 1, a reasonable choice of $c_n$ is $\min_{1\leq k\leq K}c_0\sqrt{n^{(m^*,k)}/\log p}$. With the initialization strategy 2, a reasonable choice of $c_n$ is $c_0\min_{0\leq k\leq K}\sqrt{n^{(I_k,k)}/\log p}$. In practice, a common practice is to select the tuning parameters by cross-validation.

We first show the convergence rate of the pooled transfer learning estimator $\bbeta$ in the following lemma. 

\begin{lemma}[Convergence rate of the global transfer learning estimator]
\label{lem-global}
Assume Conditions \ref{cond1} and \ref{cond2} hold. 
Assume that $h\leq s$, $N^{(0)}\log p\leq \min_{1\leq k\leq K}(N^{(k)})^2$, and\\ $\max_{0\leq k\leq K}s\log p/N^{(k)}+hs\log p/N^{(0)}=o(1)$. Then for $\hat{\bbeta}$ defined in (\ref{global3}),	for some constant $c_1\geq 1$.
	\begin{align*}
	&\sup_{\Theta(s,h)}\P\left(\|\hat{\bbeta}-\bbeta\|_2^2\gtrsim \frac{s\log p}{N}+\frac{h\log p}{N^{(0)}}\right)\leq \exp(-c_1\log p).
\end{align*}
\end{lemma}
\begin{remark}
The convergence rate of $\hat{\bbeta}$ is minimax optimal in $\ell_2$-norm in the parameter space $\Theta(s,h)$ given that $\max_{1\leq k\leq K}s\sqrt{\log p/N^{(k)}}+h\sqrt{\log p/N^{(0)}}=o(1)$ according to \citet{LZCL21}.
\end{remark}

Lemma \ref{lem-global} demonstrates that the pooled estimator $\hat{\bbeta}$ has optimal rates under mild conditions. Its convergence rate is faster than the target-only minimax rate $s\log p/N^{(0)}$ when $N^{(0)}\ll N$ and $h\ll s$. 
The sample size condition of Lemma \ref{lem-global} is relatively mild. First, $N\geq KN^{(0)}$ is easily satisfied as our target population is underrepresented. The condition that $s=o(\min_{0\leq k\leq K}N^{(k)}/\log p)$ suggests that it is beneficial to exclude too-small samples as source data. The condition that $hs=o(N^{(0)}/\log p)$ and $h\leq s$ requires that the similarity among different populations is sufficiently high.
In practice,  this assumption can be violated. In this case, Corollary \ref{remark3} shows the
aggregation step we discussed in Section \ref*{sec2-2}  prevent negative transfer and guarantee that the estimation error is no worse than only using the target data. 

Lemma \ref{lem-global} also shows the benefits of our two data integration strategies. When we integrate data across multiple sites, $N^{(k)}$ becomes larger, which relaxes the sparsity conditions and improves the convergence rate. On the other hand, when we incorporate data from diverse populations,  the total sample size $N$ is increased which also improves the convergence rate. 

\subsection{Convergence rate of Algorithm \ref{alg-ms}}
\label{sec3-1}
In this subsection, we first provide in Theorem \ref{thm1-ms} a general conclusion which describes how the convergence rate of Algorithm \ref{alg-ms} relies on the initial values. We then provide the convergence rates of  Algorithm \ref{alg-ms} under initialization Strategies 1 and 2 in Corollaries \ref{cor1-ms} and \ref{cor2-ms}, respectively. 

\begin{theorem}[Error contraction of Algorithm \ref{alg-ms}]
	\label{thm1-ms}
	Assume Conditions \ref{cond1} and \ref{cond2} and the true parameters are in the parameter space $\Theta(s,h)$. 
	Assume that $h\leq s\leq c\sqrt{N^{(0)}}$ and $\max_{1\leq k\leq K}s\log p/N^{(k)}+hs\log p/N^{(0)}=o(1)$.
If event $E_0$ in (A.14) holds for the initial estimators $\hat{\bbeta}_0$ and $\{\hat{\bw}_0^{(k)}\}_{k=1}^K$, then  with probability at least $1-\exp(-c_2\log p)$, for any finite $t\geq 1$,
	\begin{align}
	&\|\hat{\bbeta}_t-\bbeta\|_2^2\lesssim \frac{s\log p}{N}+\frac{h\log p}{N^{(0)}}+\left(\max_{1\leq k\leq K}\|\hat{\bw}^{(k)}_{0}-\bw^{(k)}\|_2+\|\hat{\bbeta}_0-\bbeta\|_2\right)^{4t}.\label{re1-beta}
	\end{align}
\end{theorem}
Theorem \ref{thm1-ms} establishes the convergence rate of $\hat{\bbeta}_t$ under certain conditions on the initializations. As the conditions in $E_0$ guarantee that $\|\hat{\bw}_0^{(k)}-\bw\|_2=o(1)$ for all $1\le k\le K$ and $\|\hat{\bbeta}^{(0)}-\bbeta\|_2=o(1)$, $\hat{\bw}^{(k)}_t$ and $\hat{\bbeta}_t$ converge to $\bw^{(k)}$ and $\bbeta$ in $\ell_2$-norm, respectively. 
For large enough $t$, the convergence rate of $\hat{\bbeta}_t$ is ${s\log p}/{N}+{h\log p}/{N^{(0)}}$, which is the minimax rate for estimating $\bbeta$ in $\Theta(s,h)$.  Hence, the proposed distributed estimators converge to the global minimax estimators.  With proper initialization, the smallest $t$ satisfying
$\left(\max_{1\leq k\leq K}\|\hat{\bw}^{(k)}_{0}-\bw^{(k)}\|_2+\|\hat{\bbeta}_0-\bbeta\|_2\right)^{4t} \le{s\log p}/{N}+{h\log p}/{N^{(0)}}$
may be very small. Detailed analysis based on the initialization strategies proposed in Section \ref{sec2-4} are provided in the sequel. 


Comparing Theorem \ref{thm1-ms} with Lemma \ref{lem-global}, we see some important trade-offs in federated learning. First, the larger estimation error of $\hat{\bbeta}_t$ with small $t$ in comparison to the pooled version $\hat{\bbeta}$ is a consequence of leverage summary information rather than the individual data.
 Second, while the accuracy of $\hat{\bbeta}_t$ improves as $t$ increases, the communication cost also increases. A balance between communication efficiency and estimation accuracy need to determined based on the practical constraints.

To better understand the convergence rate, we investigate the initialization strategies proposed in Section \ref{sec2-4}.  Under the single-site strategy, we have the following conclusion.

\begin{corollary}[Convergence rate of Algorithm \ref{alg-ms} with single-site initialization]
	\label{cor1-ms}
	We compute $\hat{\bbeta}_0$ and $\hat{\bw}_0^{(k)}$ via (\ref{global1})-(\ref{global3}) based on the individual data at site $m^*$. 
	Assume Conditions \ref{cond1} and \ref{cond2} and the true parameters are in the parameter space $\Theta(s,h)$. Assume that $N\geq KN^{(0)}$, $N^{(m^*)}\geq Kn^{(m^*,0)}$, $h\leq s$, and $\max_{1\leq k\leq K}s^2\log p/n^{(m^*,k)}+sh\log p/n^{(m^*,0)}=o(1)$.
Then with probability at least $1-\exp(-c_2\log p)$,  for any fixed $T\geq 1$,
	\begin{align*}
	&\|\hat{\bbeta}_T-\bbeta\|_2^2\leq \frac{s\log p}{N}+\frac{h\log p}{N^{(0)}}+\left\{\frac{s\log p}{N^{(m^*)}}+\frac{h\log p}{n^{(m^*,0)}}\right\}^{2T}.
	\end{align*}
\end{corollary}

Corollary \ref{cor1-ms} uses the result that $\|\hat{\bw}_0^{(k)}-\bw^{(k)}\|_2^2=O_P((s+h)\log p/n^{(m^*,k)})$ and $\|\hat{\bbeta}_0-\bbeta\|_2^2=O_P(s\log p/N^{(m^*)}+h\log p/n^{(m^*,0)})$ under the current conditions. We see that after $O(\ln N/\ln N^{(m^*)}+\ln N^{(0)}/\ln n^{(m^*,0)})$ number of iterations, $\hat{\bbeta}_t$ has the same convergence rate as the global estimator $\hat{\bbeta}$. If $N\asymp (N^{(m^*)})^{\alpha}$ and $N^{(0)}\asymp (n^{(m^*,0)})^{\alpha'}$ for some finite $\alpha$ and $\alpha'$, then only constant number of iterations are needed.


Next, we study the performance of Algorithm \ref{alg-ms} when using multi-site initialization strategy. For simplicity, we study the case where  $n^{(I_k,k)}\asymp N^{(I_k)}$. In other words, $w^{(k)}$ is initialized based on only the data from the $k$-th population in site $I_k$, i.e.,
\begin{align}
\hat{\bbeta}_0&=\argmin_{b\in\R^p}\{\frac{1}{n^{(I_0,0)}}L^{(I_0,0)}(\bb)+\lam_{\beta,0}\|\bb\|_1\},\nonumber\\
   \hat{\bw}_0^{(k)}&=\argmin_{b\in\R^p}\{\frac{1}{n^{(I_k,k)}}L^{(I_k,k)}(\bb)+\lam_{0}^{(k)}\|\bb\|_1\}, ~k=1,\dots,K.\label{bw-init}
\end{align}
\begin{corollary}[Convergence rate of Algorithm \ref{alg-ms} with multi-site initialization]
	\label{cor2-ms}
	We compute $\hat{\bbeta}_0$ and $\hat{\bw}_0^{(k)}$ based on (\ref{bw-init}) with site $I_k$ for $k=0,\dots,K$. We take $\lam_{\beta,0}=c_1\sqrt{\log p/n^{(I_0,0)}}$ and $\lam_{0}^{(k)}=c_1\sqrt{\log p/n^{(I_k,0)}}$ with some large enough constant $c_1$.
	Assume Conditions \ref{cond1} and \ref{cond2} and the true parameters are in the parameter space $\Theta(s,h)$. Assume that $N\geq KN^{(0)}$, $h\leq s$, and $\max_{1\leq k\leq K}s^2\log p/n^{(I_k,k)}+sh\log p/N^{(I_0)}=o(1)$.
Then with probability at least $1-\exp(-c_2\log p)$,  for any fixed $t\geq 1$,
	\begin{align*}
	&\|\hat{\bbeta}_t-\bbeta\|_2^2\leq \frac{s\log p}{N}+\frac{h\log p}{N^{(0)}}+\min_{0\leq k\leq K}\left\{\frac{s\log p}{n^{(I_k,k)}}\right\}^{2t}.
	\end{align*}
\end{corollary}
Corollary \ref{cor2-ms} uses the result that $\|\hat{\bbeta}_0-\bbeta\|_2^2=O_P(s\log p/n^{(I_0,0)})$ and $\|\hat{\bw}^{(k)}-\bw^{(k)}\|_2^2=O_P(s\log p/n^{(I_k,k)})$ in the current setting. In this case, we see that after $O(\max_{0\leq k\leq K}\ln N/\ln n^{(I_k,k)})$ number of iterations, $\hat{\bbeta}_t$ has the same convergence rate as the global estimator $\hat{\beta}$.

With the above analyses, we can evaluate the convergence rate for the proposed estimator obtained after the aggregation step.

\begin{corollary} \label{remark3}{(The effect of aggregation).}
Assume Conditions \ref{cond1} - \ref{cond2} hold. We show that with probability at least $1-\exp(-c_1\log p)-\exp(-c_2t)$,
	\[
	\|\hat{\bbeta}^{agg}-\bbeta\|_2^2\leq c_3\min\{\|\hat{\bbeta}_T-\bbeta\|_2^2,\|\hat{\bbeta}_T^{(tar)}-\bbeta\|_2^2\}+\frac{c_4t}{n^{(m^*,0)}}.
	\]
\end{corollary}
Through aggregation, we achieve an estimator whose estimation performance is comparable to the better performance of target-only $\hat{\bbeta}_T^{(tar)}$ and transfer learning $\hat{\bbeta}_T$. 

\subsection{Convergence rate of Algorithm \ref{alg-local}}
\label{sec3-2}
In this section, we provide theoretical guarantees for Algorithm \ref{alg-local}, which leverages local Hessian and only transmits first-order information across sites. As we discussed before, it relies on the homogeneity assumption on the distribution of $\bx^{(m,k)}$ for $m=1,\dots,M$ at each given $k$.
\begin{condition}[Homogeneous covariates]
\label{cond-homo}
Assume that $\{\bx_i\}_{i\in\mN^{(m,k)}}$ and $\{\bx_i\}_{i\in\mN^{(m',k)}}$ are identically distributed for any $0\leq k\leq K$ and $1\leq m,m'\leq M$.
\end{condition}

To simplify the theoretical result, we focus on the case $K=1$. That is, only one source population is in use. The more general case, where $K$ can be any finite integer, can be analyzed similarly but the results are harder to interpret.

In the next theorem, we analyze the error contraction behavior of Algorithm \ref{alg-local}.
\begin{theorem}[Error contraction of Algorithm \ref{alg-local}]
\label{thm-local}
Assume Conditions \ref{cond1}, \ref{cond2}, Condition \ref{cond-homo} and true parameters are in $\Theta(s,h)$. Assume that $h\leq s$, $\min_{1\leq k\leq K}n^{(m^*,k)}\geq n^{(m^*,0)}$, $\max_{0\leq k\leq K}s^2\log p/n^{(m^*,k)}=o(1)$. Suppose that event $E_0'$ in (B.1) holds and tuning parameters satisfy (B.2). Then with probability at least $1-\exp(-c_1\log p)$, it holds that
\[
\|\hat{\bbeta}_T-\bbeta\|_2^2\lesssim \frac{s\log p}{N}+\frac{h\log p}{N^{(0)}}+ \big(\max_{1\leq k\leq K}s(\lam_0^{(k)})^2+\|\hat{\bbeta}_0^{(0)}-\bbeta\|_2^2\big)(\frac{s^2\log p}{n^{(m^*,0)}})^{T}.
\]
\end{theorem}
Theorem \ref{thm-local} provides the error contraction analysis of Algorithm \ref{alg-local}. The event $E_0'$ in (B.1) assumes the consistency of initial estimators and specifies the tuning parameters. In fact, the tuning parameters of Algorithm \ref{alg-local} depend on the convergence rates of initial estimators and hence depend on the unknown $s$ and $h$. In the single-task first-order method with $\ell_1$-regularization (Section 3.2 in \citet{jordan2018communication}), the tuning parameters also depends on unknown parameters.
In practice, specifying these tuning parameters can be challenging and the practical performance can be less accurate without proper tuning.

In the following two corollaries, we provide convergence rate analysis of Algorithm \ref{alg-local} under two initializations proposed in Section \ref{sec2-4}.

\begin{corollary}[Convergence rate of Algorithm \ref{alg-local} with single-site initialization]
\label{cor-local}
Assume Conditions \ref{cond1}, \ref{cond2}, and Condition \ref{cond-homo}. Assume that $h\leq s$, $\min_{1\leq k\leq K}n^{(m^*,k)}\geq n^{(m^*,0)}$, $\max_{0\leq k\leq K}s^2\log p/n^{(m^*,k)}=o(1)$. Suppose that tuning parameters satisfy (B.2). Then with probability at least $1-\exp(-c_1\log p)$, it holds that for any finite $T\geq 1$,
\begin{align*}
\|\hat{\bbeta}_T-\bbeta\|_2^2\lesssim  \frac{s\log p}{N}+\frac{h\log p}{N^{(0)}}+ \left(\frac{s\log p}{\min_{1\leq k\leq K}n^{(m^*,k)}}+\frac{h\log p}{n^{(m^*,0)}}\right)(\frac{s^2\log p}{n^{(m^*,0)}})^{T}.
\end{align*}
\end{corollary}
For $\hat{\bbeta}_T$ obtained from Algorithm \ref{alg-local}, we see that it requires $O(\ln N/\ln n^{(m^*,0)})$ iterations to achieve the minimax optimal rate.
We now compare the theoretical performance of Algorithm \ref{alg-ms} and Algorithm \ref{alg-local} with single-site initialization.
In comparison to the upper bound derived in Corollary \ref{cor1-ms}, we see that the convergence rate of Algorithm \ref{alg-ms} is always no worse than the rate of Algorithm \ref{alg-local} for any given $T$. Hence, to reach comparable performance, the local Hessian algorithm requires more iterations and hence more rounds of communication. This implies that transmitting Hessian matrices not only allows heterogeneous covariates but can accelerate the convergence of federated estimators.

\begin{corollary}[Convergence rate of Algorithm \ref{alg-local} with multi-site initialization]
\label{cor-local2}
Assume Conditions \ref{cond1}, \ref{cond2}, and Condition \ref{cond-homo}. Assume that $h\leq s$, $\min_{1\leq k\leq K}n^{(m^*,k)}\geq n^{(m^*,0)}$, $\max_{0\leq k\leq K}s^2\log p/n^{(m^*,k)}=o(1)$. Suppose that tuning parameters satisfy (B.2). Then with probability at least $1-\exp(-c_1\log p)$, it holds that
\begin{align*}
\|\hat{\bbeta}_T-\bbeta\|_2^2\lesssim\frac{s\log p}{N}+\frac{h\log p}{N^{(0)}}+ \frac{s\log p}{\min_{0\leq k\leq K}n^{(I_k,k)}}(\frac{s^2\log p}{n^{(m^*,0)}})^{T}.
\end{align*}
\end{corollary}

In Corollary \ref{cor-local2}, we provide the convergence rate of Algorithm \ref{alg-local} with multi-site initialization. In comparison to Corollary \ref{cor2-ms}, the local Hessian algorithm has slower convergence rate at any given $T$. It  requires $O(\ln N/\ln n^{(m^*,0)})$ iterations to achieve the minimax optimal rate.

\section{Simulation studies}
\label{sec-simu}
In this section, we evaluate the performance of the proposed methods in terms of both estimation and prediction accuracy using a logistic regression model. Motivated from our real data application which is introduced in Section~\ref{sec-data}, we generate data to mimic polygenic risk prediction in a federated network with  $M = 5$ sites. In  site $m \in \{1, \dots M\}$, we have $n^{(m,0)} = 400$ samples from the target population and $n^{(m,1)}  = 2000$ samples from a source population. We set the dimension of $\bx_i$ to be $p = 2000$ and generate $\bx_i$ to mimic the genotype data from different ancestry groups. More specifically, for data in the source population, we first generate $p$-dimensional multivariate Gaussian vector $\bm z_i$ with mean $\bm 0$ and covariance matrix $\Sigma_{1}$. We choose $\Sigma_{1}$ to be a block-wise matrix with $20$ blocks each has dimension $100\times 100$. We set the all the $20$ blocks to be the same, denoted by $B_{1}$, where ${B_{1, ij}} = {0.5}^{|i-j|}$. We then randomly generate minor allele frequencies for the $p$ genetic variants from $U(0, 0.5)$. Then we obtain $\bx_i$ by  categorize each $\bm z_i$ into $0, 1$ and $2$ based on the corresponding minor allele frequencies. For the target data, we follow the same procedure with $\Sigma_{1}$ replaced by $\Sigma_{0}$, which has $100$ blocks each with dimension $50\times 50$. We set the block to be ${B_{0, ij}} = {0.3}^{|i-j|}$. 

For each subject, we generate the binary outcome variable through a logistic regression model 
\[
\text{logit}(\E \{y_{i}|\bx_i\})= \bx_{i}^{\intercal}\bb_i,
\] 
where $\text{logit}(t) = \log \{x/(1-x)\}$. 
The regression coefficients $\bb_i = \bbeta$ is subject $i$ is from the target population, otherwise $\bb_i = \bw$.  The regression-coefficient $\bbeta$ has $s = 100$ non-zero entries which are generated from $U(-0.5,0.5)$, and  $\bw$ is generated from the following two settings:

(S1) $w_j  =\beta_j + \Delta\mathbb{I} (j \in H)$ where $H$ is a random subset of $[p]$ with $|H| = h$. We take $h \in \{10, 20, 30\}$, and $\Delta \in \{0.5, 1, 1.5\}$, which is corresponding to the setting that  a small number of genetic variants have relatively large differences in effect sizes across populations. 

(S2) $w_j  =\beta_j + \Delta_j\mathbb{I} (j \in H)$ where $H$ is a random subset  with $|H| = h$. We generate  $\Delta_j \sim_{i.i.d.} N(0,\Delta)$. We take $h \in \{50, 100, 150\}$, and $\Delta \in \{1/6, 1/3, 1/2\}$, which is corresponding to the setting that  a large number of genetic variants have relatively small differences in effect sizes across populations. 

We compare a list of methods including (1) federated learning based on all the target data ({\it target-only}); (2) federated learning based on all the source data ({\it source-only}); (3) federated learning based on all data combing both the source and target ({\it combined}); (4) our proposed approach with $T=1$  ({\it proposed ($T = 1$)}); (5) our proposed approach with $T=3$  ({\it proposed ($T = 3$)}); (6) the pooled transfer learning (equations (\ref{global1})-\ref{global3}) method where data from all sites are pooled together  ({\it pooled}). The methods are evaluated based on their mean squared error (MSE) and the out-sample area under the receiver operating characteristic curve (AUC) based on a randomly generated testing sample with sample size $n = 1000$. 

As the results for MSE and AUC are relatively similar,  we present in Figures \ref{fig2} and  \ref{fig3}  the  AUC over $200$ replications under different simulation settings, and defer the MSE results to the Supplementary Material.  Under setting (S1), the level of heterogeneity are captured by both $\Delta$ and $h$, since $\Delta$ measures the absolute value of the non-zero entries of $\bdelta$, and $h$ measures the number of non-zero entries of $\bdelta$. We observe from Figure \ref{fig2} that with the increase of $\Delta$ and $h$, the prediction performance of the source-only estimator in the target population decreases. This is consistent with our intuition that higher level of heterogeneity will cause lower transferibility of the source estimator to the  target population. The estimator combining source and target data outperforms  the source-only estimator in all scenarios as it incorporates target samples.  Our proposed estimators, even with only one round of communication, achieve better performance than these benchmarks. With $T=3$, our methods can further improve the performance, especially when heterogeneity is high. Our estimators have comparable performance  to the pooled estimator when the heterogeneity is not very high. More iterations are needed to fill the gap between the proposed estimator and the pooled estimator. Figure \ref{fig3} conveys similar conclusions. We see that numerically, when $h$ is relatively large (even bigger than $s$), our methods still have improvement given the magnitude of entries are small. 
We also evaluated the effect of aggregation by comparing the proposed estimator with $\hat\bbeta$ from Algorithm 1 (see Figure S1 in the Supplementary Material). Across all scenarios, the proposed estimators with aggregation perform no worse than the estimators without aggregation. The improvement is substantial when the level of heterogeneity are moderate or large.

In sum,  the simulation study demonstrates that our methods provide improved estimation and prediction accuracy, compared to the benchmark methods. Compared to the ideal case where data are pooled together, our method with only one iteration has  comparable or slightly worst performance for most of the scenarios. When the level of heterogeneity is high, extra iterations can  improve the accuracy and fill the gap between the federated analysis and the pooled analysis. In general, we recommend the aggregation step to ensure the robustness of the results especially when the level of heterogeneity is high. 
\begin{figure}
	\caption{Comparison of AUC over $200$ replications under simulation setting (S1).}\label{fig2}
	\centering
	\includegraphics[width=15 cm]{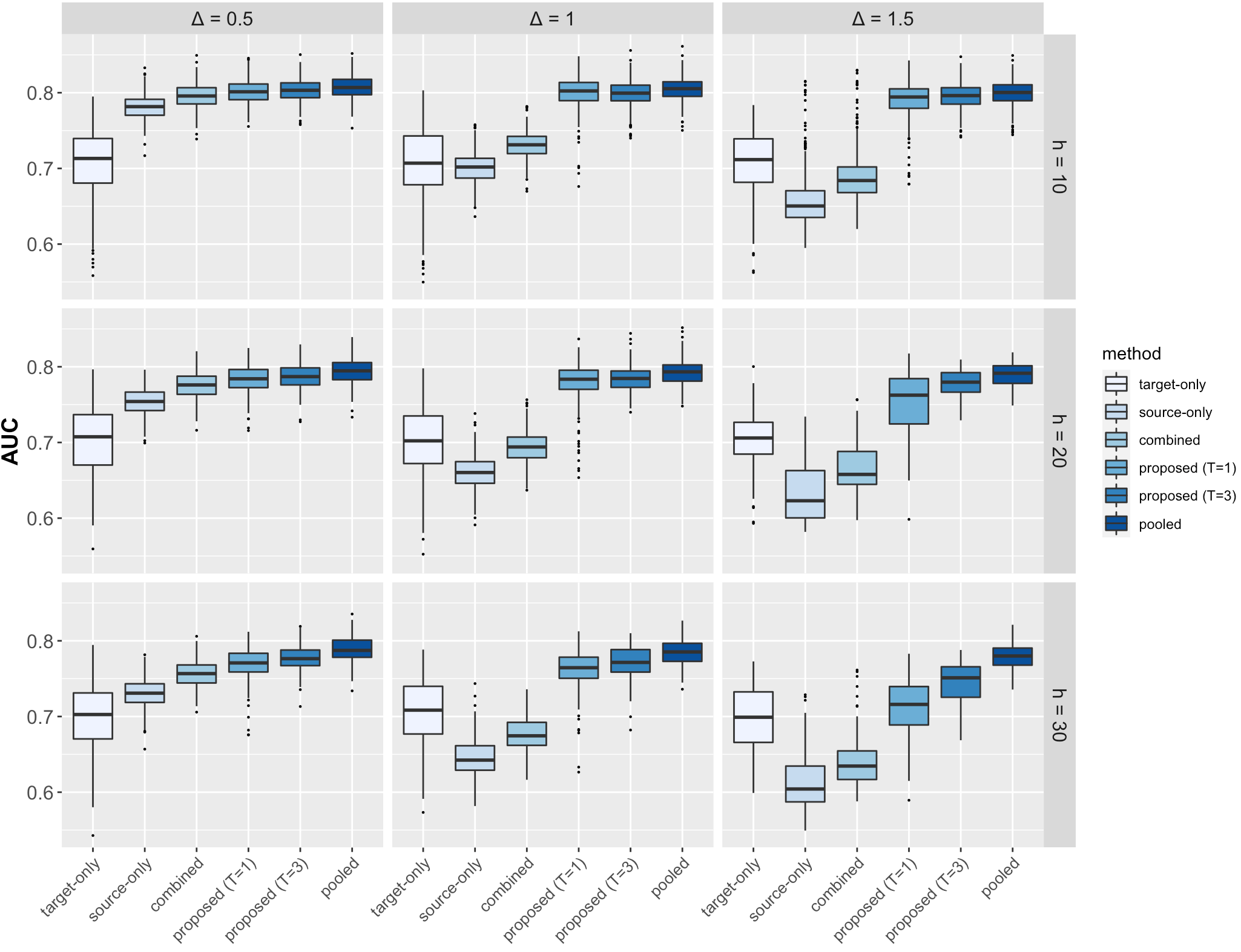}
\end{figure}

\begin{figure}
	\caption{Comparison of AUC over $200$ replications under simulation setting (S2).}\label{fig3}
	\centering
	\includegraphics[width=15 cm]{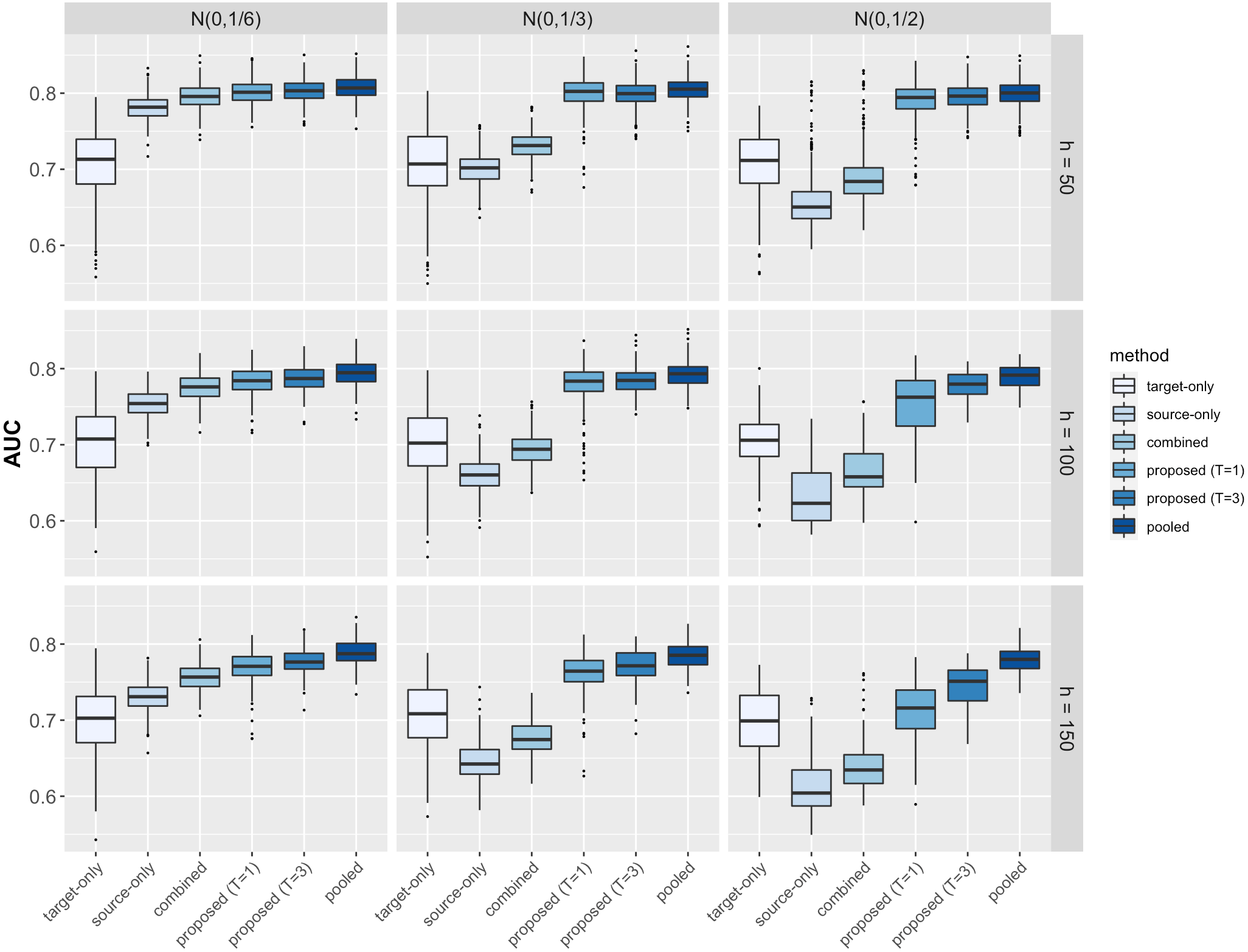}
\end{figure}


\section{Application to construct PRS for Type II diabetes using data from the eMERGE network}
\label{sec-data}
We evaluate our proposed methods using multicenter data from the eMERGE Network Imputed genome-wide association studies, where the data were collected from EHR-derived phenotypes and DNA from linked Biorepositories. The dataset contains genotype and phenotype information for 55,029 subjects from ten participating sites (Cincinnati Children's Hospital Medical Center/Boston Children's Hospital, Children's Hospital of Philadelphia, Essentia Institute of Rural Health, Marshfield Clinic Research Foundation and Pennsylvania State University, Geisinger Clinic, Group Health Cooperative/University of Washington, Mayo Clinic, Icahn School of Medicine at Mount Sinai, Northwestern University, Vanderbilt University Medical Center).  

In this application, we apply our methods to construct polygenic risk scores for Type II Diabetes, each participant was labelled as a case, a control  or unknown. In addition, we also observe the self-reported race for all the participants. The dataset contains 73\% of the participants are White, 20\% of the participants are Black or African American,  1\% are Asian and 6\% are Unknown. 
As the sample sizes for Asian participants are extremely small, we consider only two racial groups, White and African American. We treat the self-reported race as an approximation for the ancestry indicator (EA and AA). We treat AA as our target population, and EA as the source population. After removing samples with unknown disease status, we obtain in total 20,247 samples from seven participating sites. 

\begin{table}
	\begin{center}
		\begin{tabular}{ cccc }
			\hline
			\hline
			Site Name & N & \# of EA & \# of AA\\
			\hline
			Geisinger Health System & 3090 & 3081& 9\\ 
			Group Health Cooperative&    278 &    262&                               16  \\ 
			Marshfield Clinic&3980&3977& 3 \\
			Mayo Clinic&2890&2880 &10 \\
			Mount Sinai &4259  &606   &3653      \\                             
			Northwestern University &1524&1223  &301   \\         
			Vanderbilt University &4226&   2580& 1646\\
			\hline
		\end{tabular}
		\caption{Sample sizes of AA and EA populations across seven sites.}
	\end{center}
\end{table} \label{table1}

Table \ref{table1} shows the characteristics of the samples by site. We observe that participants from Geisinger, Group Health,  Marshfield and Mayo are mostly from EA population, while less than 20 subjects are from AA population. Since the sample size less than 20 may cause potential issues for privacy protection, we exclude the AA samples from Geisinger, Group Health,  Marshfield and Mayo. The three sites with substantial numbers of AA samples are Mount Sinai, Northwestern, and Vanderbilt. In our analysis, we choose Northwestern as an external testing dataset, where we do not included it in the model training. In addition, we randomly select $300$ African American samples from Vanderbilt and Northwestern, respectively, for internal testing. We filtered the single nucleotide polymorphisms (SNPs) based on Hardy–Weinberg equilibrium, missing proportion, minor allele frequency,  LD pruning, and marginal effect sizes obtained from an external large GWAS \citep{morris2012large}. The detailed filtering parameters can be found in the Supplementary Material. A total of $2017$ SNPs  passed the filtering process and are included in the prediction model. 

We applied our proposed method with number of iterations $T = 1$ and $T = 3$. As we show in the simulation study that the all-data estimator combining the target and the source is always better than the source-only estimator, we compare the proposed estimator with the  target-only estimator and the all-data estimator.  Two metrics are used to evaluate methods based on the three testing datasets from Mount Sinai,  Vanderbilt and Northwestern,  (1) the area under the receiver operating characteristic curve (AUC), and (2) odds ratio comparing the top 20\% of patients with the bottom 20\% of patients according to the PRS distribution.  The odds ratios are often used in PRS research as a metric show how well the PRS can  stratify high-risk patients vs low risk patients. To account for sampling variation, we repeat the evaluation $20$ times and within each replication, we randomly choose the $300$ testing samples from Vanderbilt and Mount Sinai. The evaluations are done separately on the three testing sets. 


\begin{figure}
	\caption{Comparisons of AUC and odds ratio of compared methods across three testing datasets}\label{data_auc}
	\centering
	\includegraphics[width=16 cm]{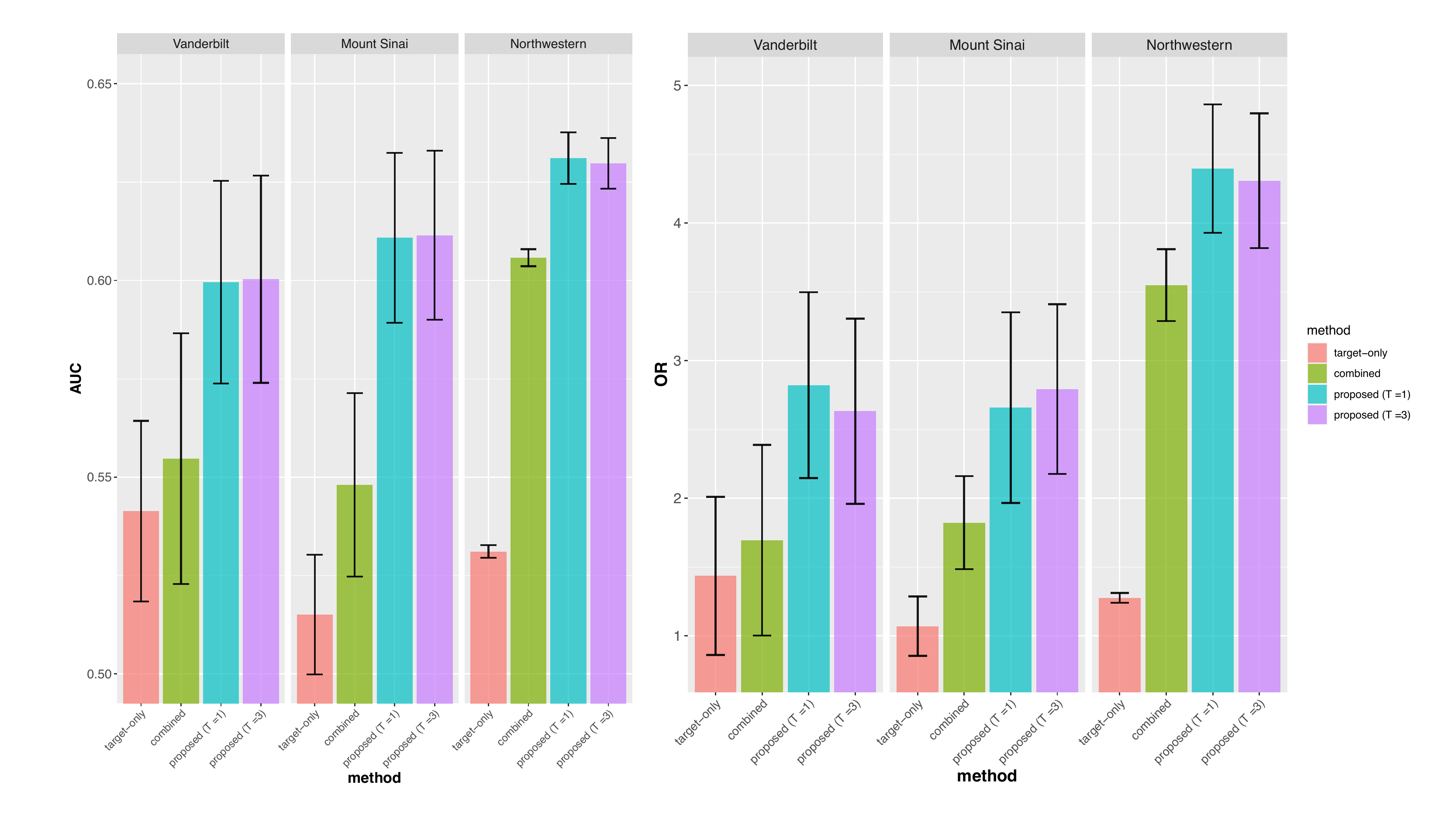}
\end{figure}

Figures \ref{data_auc} presents the performance of the compared methods. We observe that the target-only estimator performs generally poorly across the three testing datasets, with AUC ranges from $0.52$ to $0.54$. The combined estimator, performs better than the target-only estimator, having AUC around $0.55$ to $0.58$. The proposed estimator  $T=1$ improves the AUC to $0.60-0.63$. The proposed estimator with two extra iterations $T=3$ does not perform better than $T=1$ in this application. The AUCs are nearly the same the $T=1$. When comparing the odds ratio, we obtained similar conclusions that the proposed estimator with $T=1$ can best separate the high risk and low risk groups. The variation of performance across different testing data implies potential heterogeneity across sites. 

In sum, this application demonstrate the feasibility and the promise of our methods to be implemented in large genomics network for construction of polygenic risk scores. It may also be applied in other use case, such as EHR phenotyping, or EHR-based risk profiling using multicenter data.

\section{Discussion}
In this paper, we propose federated transfer learning methods to improve the performance of estimation and risk prediction in underrepresented populations. Our methods allow incorporating data from diverse populations that are stored at different institutions. We provide theoretical analysis and numerical experiments to demonstrate that our methods provide more accurate estimators for the underrepresented population compared with benchmark methods. And through a limited number of iterative communications across sites, our methods can achieve similar accuracy as the pooled analysis which requires directly sharing individual-level data. We obtained promising results from a real application to eMERGE network for constructing polygenic risk prediction models for Type II diabetes in AA population, which demonstrates the feasibility of applying our methods to large clinical/genomics consortia for risk profiling and prediction in diverse populations.

Although our methods are based on high-dimensional GLMs using  $L$1 penalty, it can be easily extended to $L2$ penalty or elastic net type of penalties, depending on whether the difference of regression parameters between a source and the target population is sparse or nearly sparse. In our software package, we allow different types of penalty functions which can be chosen by cross validation.

We account for population-level heterogeneity by allowing both the conditional distribution $f(y|\bx)$ and the marginal distribution $f(\bx)$ to be different across sites. We use an aggregation method to improve the robustness of our methods to the level of heterogeneity. To account for site-level heterogeneity, our methods  allow $f(y|\bx)$ to vary across sites, while  $f(y|\bx)$ is assumed to be shared across sites given a specific population. In practice, there might still be site-level heterogeneity that cause differences in $f(y|\bx)$ and robust methods to account for such heterogeneity need to be incorporated, which we will  consider in our future work.

Our proposed methods improve the fairness of statistical models by reducing the gap of estimation accuracy across populations due to lack of representation. 
Our theoretical conclusion provides insights for future data collection, as it reveals how the level of heterogeneity impacts the accuracy gain, and the sample size from the target population needed to achieve comparable accuracy as the source populations. This is  different from methods that impose constraints on model fitting to ensure that the prediction accuracy of an algorithm has to be at the same level across different populations \citep{mehrabi2021survey}.  Since we are focusing on improving the performance of models in an underrepresented population, our work cannot guarantee complete fairness in prediction accuracy across all groups. In the future, we can incorporate fairness corrections and constraints into our framework, and in the meantime consider a wide variety of models in longitudinal and survival analysis, causal inference, to advance algorithmic fairness in precision medicine.

\bibliographystyle{chicago}
\bibliography{Distributed-TL2}

\appendix

\end{document}